# A Reactive Tabu Search Algorithm for Stimuli Generation in Psycholinguistics

Alejandro Chinea
Department of Cognitive Neuroscience
University of La Laguna
Campus de Guajara s/n, 38205- Tenerife, SPAIN
alchinea@ull.es

**ABSTRACT**

*The generation of meaningless "words" matching certain statistical and/or linguistic criteria is frequently needed for experimental purposes in Psycholinguistics. Such stimuli receive the name of pseudowords or nonwords in the Cognitive Neuroscience literature. The process for building nonwords sometimes has to be based on linguistic units such as syllables or morphemes, resulting in a numerical explosion of combinations when the size of the nonwords is increased. In this paper, a reactive tabu search scheme is proposed to generate nonwords of variable size. The approach builds pseudowords by using a modified Metaheuristic algorithm based on a local search procedure enhanced by a feedback-based scheme. Experimental results show that the new algorithm is a practical and effective tool for nonword generation.*

## 1. INTRODUCTION

In the last few years there has been a great deal of cognitive neuroscience research into how language is processed, acquired, comprehended and produced by the human brain [1][2]. Two major tools in this research area are computational models and laboratory experiments in which language features are manipulated. Computational models try to simulate how language information is processed, while psycholinguistics experiments record behavioral responses such as reaction times, or the electrophysiological or haemodynamic responses of human subjects to specific linguistic stimuli. Thus, the experiments test the predictions of the computational models with the aim of understanding the representation and processing of language components in the human brain.

In order to empirically test hypotheses and models, cognitive neuroscience researchers have frequently faced the problem of generating appropriate linguistic stimuli for their experiments. This involves, in some cases, searching for words with well-defined statistical and/or linguistic properties (e.g., words within specific ranges of printed frequency, syllable frequency, number of neighbors and so forth), and/or nonwords (i.e, stimuli that resemble a word but are not part of the words of a particular language; for instance, "pint" is an English word, but "pont" is not) also with special properties. It is important to stress that large numbers of nonwords are needed in many experimental paradigms. For instance, the lexical decision task, widely used in cognitive neuroscience implies the use of words and nonwords that participants have to discriminate between. There are some very recent tools for generating pseudowords in languages such as English [3][4]. However, to our knowledge there are no general-purpose tools or algorithms for polysyllabic nonword generation which are able to generate nonwords for any given language and are subject to any given criteria. Given the variety of demands of most experimental paradigms, it is surprising that many of the stimuli are at present practically generated by hand from linguistic databases or employing very primitive tools that are time consuming. This is due to several factors. Firstly, that the criteria of word-generation in a single experiment can be varied. In addition, most European languages are made up of thousands of different syllables as well as morphemes. It is not, then, difficult to deduce that as word-size increases, measured as the number of word granules, the number of possible combinations increases exponentially. Therefore, nonword generation, if not automated, leads to a cumbersome and error-prone task that also turns into a combinatorial search problem when the size of the nonwords is to be considered.

In this paper, we propose an algorithm to automate the process of nonword generation for any given criteria. The algorithm is based on a Reactive Tabu Search approach [5][6]. This framework of algorithms is characterized by the introduction of feedback schemes in heuristics for discrete optimization problems. These are intensively memory-based and they present robust search capabilities which effectively seek solutions from a vast search space at reasonable computational cost. Most of the present variants are utilized for both combinatorial optimization problems and real-valued function optimization [7][8]. This paper explores the possibility of using a Reactive Tabu Search scheme for a combinatorial search task. More specifically, we are interested in finding the highest number of solutions, as there is no single globally-optimal solution to the problem, in contrast to a classic combinatorial optimization problem.

The rest of this paper is organized as follows: In the next section, the problem we address is presented. To emphasize the characteristics of the problem a brief analysis of complexity is made, reviewing some aspects of combinatorial optimization. Section 3 is a formal description of the problem and the approach proposed: The adaptation of a Reactive Tabu Search scheme to a combinatorial search task. Section 4 focusses on the application of the proposed scheme to a specific case study. The most important parts of the algorithm are sketched in this section. The experimental results are covered in section 5, with some implementation and performance details. Finally, section 6 provides a summary of the present study and some concluding remarks.

## 2. PROBLEM STATEMENT

Let us consider the problem of generating nonwords using permutations of a basic word unit µ in any given language. In general, a basic word unit is defined as the granules in which a given word can be decomposed in terms of a predefined linguistic property. Actually, most commonly used word granules are syllables and morphemes, although others could also be used such as bigrams or trigrams (sets of two and three consecutive letters within a word respectively). Let us suppose that the basic unit µ of the language under consideration has a cardinality equal to $\lambda$, denoting as cardinality the range of different values a word unit can hold. Therefore, it can be deduced that in the process of generating nonwords of size d, measured as the number of word-units used to compose the nonwords, the number of possible combinations is $\lambda^d$. More specifically, the number of combinations increases exponentially with d. Hence, the associated search complexity varies considerably depending on the language under consideration and the search criteria to be used.

Most Romance languages include multiple syllable words, for instance, in Spanish and Portuguese there are words of up to ten syllables, or in French up to nine syllables. Furthermore, non-Romance languages like German, which is an agglutinative language (some words are composed by the union of more words) also have words with a high number of syllables and morphemes. Specifically, the German language contains words with more than ten syllables. Taking into account the fact that for many of the languages, the cardinality of the word units µ is in the order of thousands (eg: There is approximately 4,000 different syllables in Spanish, and 6,000 in German), the process of nonword generation becomes a combinatorial search problem, as pointed out in the previous discussion. In principle, to deal with the above-mentioned problem, a first approach could be based on a raw search procedure by using brute-force. However, depending on the search criteria, as well as the evaluation criterion (eg: A time-consuming evaluation criterion is not feasible even for reasonably sized low dimensional search spaces) this method cannot provide a satisfactory number of solutions in all cases. Furthermore, its convergence to a useful number of solutions is not always guaranteed. In addition, the computational resources required by this approach could be prohibitive in many cases. A promising way to solve this problem is to adapt a combinatorial optimization algorithm to a merely combinatorial search task. Metaheuristic algorithms offer a good alternative in this line. Here, a Reactive Tabu Search (henceforth RS) scheme is considered in the following discussion.

The general structure of a RS algorithm consists in a local search heuristic based on a specific neighborhood, complemented with a memory-based mechanism designed to avoid cycles. The RS framework employs a simple form of reinforcement learning as it is intensively history-based, therefore implementing a sort of knowledge acquisition via generalizing upon experiences of prior functioning. The most important parts in a RS algorithm are: the objective function to be optimized, the chosen representation for potential solutions to the problem and the neighborhood generation procedure. However, these algorithms are designed to work for discrete optimization problems. The nonword generation process is a combinatorial search problem where the objective function is Boolean. Specifically, generated nonwords match or not the designed search criteria. Therefore, there is no single globally optimal solution to the problem. Furthermore, we are interested in finding the highest amount of nonwords using the minimal computational resources. In the following subsections the details of our approach are presented, together with a specific case study.

## 3. PROPOSED SOLUTION

Let us define the notation. Given an instance of a Combinatorial search problem, i.e., a pair *(V, C)*, where *V* is a set of feasible points with finite cardinality and *C* is a 0-1 criterion function which evaluates points in the combinatorial space: $V: C \rightarrow \{0,1\}$. The neighborhood *N(v)* associates to each point *v* a subset of *V*. A point *v* is optimal with respect to *N* if $C(v) = 1$. Moreover, any point $v \in V$ is said to accomplish the optimality criterion if and only if $C(v) = 1$. For the following discussion the neighborhood *N(v)* is defined as the set of points that can be obtained by applying to *v* a set of elementary moves *M*.

The basic RS scheme uses an iterative greedy search algorithm to bias the search toward points with lower or higher values of the objective function (depending on whether we are minimizing or maximizing the objective function respectively). In our approach, as the *C* function is Boolean, a transformation is made to convert this function into a real-valued function by using some distance function: $C^l = G(C(v))$, so now: $V: C^l \rightarrow [0,1] / \forall \ v \in V \ C^l(v) = 1 \Leftrightarrow C(v) = 1$. This relaxation is introduced to measure how far points in the feasible space are from the optimality criterion. In addition, with regards to the basic RS scheme, the neighborhood (or set of basic moves) is designed following a randomization procedure. The core idea underlying this approach is to increase the diversification and exploration capabilities of the algorithm. It is important to emphasize the fact that we are interested in finding the highest number of solutions.

At any given iteration *t* of the search, the set of moves *M* is partitioned into the set $P^{(t)}$ of the not allowed moves and the set $A^{(t)}$ of admissible moves (the moves that can be applied to the current point). In addition, $A^{(t)}$ is subdivided in two sets $A_1^{(t)}$, $A_2^{(t)}$, $A^{(t)} = A_1^{(t)} \cup A_2^{(t)}$, where $A_1^{(t)}$ contains the points matching the optimality criterion and $A_2^{(t)}$ the rest of the admissible moves. At the beginning, the search starts from an initial configuration $v^{(0)}$, which is generated randomly and all moves are admissible: $A^{(0)} = M$, $P^{(0)} = \emptyset$. At a given iteration *t*, the successor of the current point is obtained by selecting the best move $\mu(t)$ (figure 3) from the set $A^{(t)}$ following the procedure:

$$v^{(t+1)} = \mu^{(t)}(v^{(t)}) \quad \text{where} \quad \mu^{(t)} = \begin{cases} \arg\max_{\eta \in A_2^{(t)}} C^1(\eta(v^{(t)})) & |A_1^{(t)}| = 0 \\ \text{random } \eta(v^{(t)}) & |A_1^{(t)}| > 0 \\ \eta \in A_1^{(t)} \end{cases}$$

The above procedure is always true unless $A^{(t)}$ becomes empty, in that case one point is randomly selected from the set $P^{(t)}$. Unlike with the RS scheme, the aspiration criterion is not present within the function **best_move**. During the local search procedure (or neighborhood search), all the points giving the maximum value of $C^1$, which have not been previously found, are stored in a local data structure. At the end of the procedure the size of the structure is checked. Then, if the size is different from zero, a point within the data structure is randomly selected. Otherwise, from the local visited points, the one with the highest score is chosen. At the end of this procedure, all the solutions found in the neighborhood are inserted into a global container data structure *D* that guarantees the absence of duplicates. The sets of points $v^{(t)}$, obtained by the above procedure is called a trajectory. It is possible that $v^{(t+1)} = v^{(t)}$, thus, returning to the previous configuration. Several mechanisms are implemented to deal with cycles: Firstly, the inverses of the moves executed in the most recent part of the search are prohibited. Furthermore, an additional mechanism to avoid cycles is implemented by dynamically adapting the prohibition period of the moves. More specifically, the most recent iteration when each move $\mu_i$ has been applied is recorded and each configuration $v^{(t)}$, visited by the search trajectory is stored in memory with the most recent time when it was encountered. Let us introduce the functions:

- $\Lambda(\mu)$ : The last iteration when $\mu$ has been used ($\Lambda(\mu) = -\infty$ if $\mu$ has never been used)

- $\Pi(v)$ : The last iteration when *v* has been encountered ($\Pi(v) = -\infty$ if *v* has not been found or if it is not in memory).

- $\Phi(v)$ : The number of repetitions of configuration *v* in the search trajectory. At the beginning the repetition counter $\Phi(v)$ is equal to zero for all configurations.

The prohibition period is defined as the number of iterations that a move remains in the $P^{(t)}$ set. Moreover, it can be viewed as the minimum number of iterations that must separate the repetition of the same configuration on a trajectory. In the initialization part of the algorithm the prohibition period is set to a small value (e.g., $T^{(0)} \leftarrow 1$), and later on adapted by reacting to the occurrence of repetitions. As in the basic scheme, the prohibition period $T^{(t)}$ is time-dependent: a move $\mu$ is prohibited if and only if its most recent use has been at time $\tau \geq (t - T^{(t)})$. It is also important to note that solutions found during the local search procedure are not prohibited, but only the points visited throughout the trajectory generated by the algorithm.

```
function combinatorial_reactive_search

(Initialize the prohibition data structures)
t ← 0                    (iteration counter)
T⁽⁰⁾ ← 1                 (prohibition period)
t_T ← 0                  (last time prohibition period was changed)
S ← 0                    (set of often repeated configurations)
R_ave ← 1                (moving average of repetition interval)
v⁽⁰⁾ ← random            (initial configuration)
D⁽⁰⁾ ← ∅                 (data structure to keep track of solutions found)

repeat

    escape ← memory_based_reaction( v(t) )

    if ( escape == false )
        μ ← best_move  (D⁽ᵗ⁾ is implicitly changed)
        v⁽ᵗ⁺¹⁾ = μ( v⁽ᵗ⁾ )
        Λ(μ) ← t  (A⁽ᵗ⁾ and P⁽ᵗ⁾ are implicitly changed)
        t ← t+1
    else
        diversify_search

until ||D|| is acceptable or maximum number of iterations reached

endfunction
```

Figure 1: Main structure of the algorithm

The skeleton of the algorithm is depicted in figure 1, we have named the proposed algorithm *Combinatorial Reactive Search* (CRS hereafter). After the initialization of the algorithm, the main loop is executed until a satisfactory number of solutions is found or a limiting number of iterations is reached. In the first statement of the loop, the current configuration is compared with the previously visited points stored in memory by calling the function **memory_based_reaction** (Figure 2), that returns a binary value. When the return condition is false (*escape = 0*), the next move is selected by calling the function **best_move** (Figure 3), in the other case (*escape = 1*) the algorithm enters into a diversification phase based on a random walk by calling the function **diversify_search** (see Figure 3).

Limited cycles and confinements in limited portions of the search space are discouraged by the reactive mechanisms defined by the algorithm that modify the discrete dynamical system defined by the trajectory. The reaction is based on the past history of the search and it causes possible changes of $T^{(t)}$ or the activation of a

```
function memory_based_reaction( v )

    if Π(v) > -∞ then
        Find the cycle length, update last_time and repetitions
        R ← t - Π(v)
        Φ(v) ← Φ(v)+1
        if Φ(v) > REP then
            S ← S ∪ v (v is added to the set of often repeated
                       configurations)
            if ||S|| > CHAOS then
                S ← ∅
                return true
            endif
        endif
        if R < R_max then
            R_ave ← 0.1 x R + 0.9 x R_ave
            T^(t+1) ← T^(t) x INCREASE
            t_T ← t
        endif
    else
        (if the configuration is not found, install it and set)
        Π(v) ← t
        Φ(v) ← 1
    endif
    if (t - t_T) > R_ave then
        T^(t+1) ← Max ( T^(t) x DECREASE, 1 )
        t_T ← t
    endif

    return false

endfunction
```

Figure 2: The function memory_based_reaction

diversifying phase. Short cycles are avoided by modifying $T^{(t)}$. By inspecting the function **memory_based_reaction** it can be deduced that the current configuration $v$ is compared with the configurations visited previously and stored in memory. If $v$ is found, its last visit time $\Pi(v)$ and repetition counter $\Phi(v)$ are updated. Then, if its repetitions are greater than a threshold *REP*, $v$ is included into the set *S*, and if the size $||S||$ is greater than the threshold *CHAOS*, the function returns immediately the value *true*. If the repetition interval is sufficiently short ($R < R_{max}$) cycles can be discouraged by increasing $T^{(t)}$ as in the basic RS scheme: $T^{(t+1)} \leftarrow T^{(t)}$ x INCREASE . The value of $R_{max}$ depends on the problem representation (encoding of the solutions) together with the local search procedure utilized. Similarly, if $v$ is not found, it is stored in memory, the most recent time it was encountered is set to the current time ($\Pi(v) \leftarrow t$) and its repetition counter is set to one ($\Phi(v) \leftarrow 1$).

When the reaction that modifies $T^{(t)}$ is not sufficient to guarantee that the trajectory is not confined in a limited portion of the search space, the search dynamics enter a phase of random walk specified by the function **diversify_search**. Specifically, when this phase begins the memory structure is cleaned, although $R_{ave}$ and $T^{(t)}$ are not changed. The number of random steps is proportional to $R_{ave}$. Afterwards, the most recent random steps are prohibited to avoid the dynamical system from returning to the old region.

```
function best_move

    M ← neighborhood_generation( v )
    if || A^(t) || = 0 then
        μ ← arg max C^1 ( η ( v^(t) ) )   (η ∈ P^(t) )
        T^(t+1) ← T^(t) x DECREASE
        t_T ← t
    else
        if || A_1^(t) || = 0 then
            μ ← arg max C^1 ( η ( v^(t) ) )   (η ∈ A_2^(t) )
        else
            μ ← random η ( v^(t) )   (η ∈ A_1^(t))
            D^(t) ← A_1^(t)
        endif
    endif

    return μ

endfunction

function diversify_search

    ( Cleaning memory structures Π and Φ )
    Π ← 0
    Φ ← 0
    ξ ← { Min ( 1+ R_ave/2, ||M|| ) }
    repeat
        σ ← random (generate a random configuration)
        v^(t+1) ← σ
        Λ (σ) ← t   ( A^(t) and P^(t) are implicitly changed )
        t ← t+1
    until the number of loop repetitions equal to ξ

endfunction
```

Figure 3: The function best_move and diversify_search

## 4. A PRACTICAL CASE STUDY

The general ideas behind the approach presented in the previous section are illustrated in the following discussion through two specific practical applications. Let us first, however, introduce some definitions:

(1) Definition of bigram:

A bigram is defined as two consecutive characters within a word (eg: The word "table" admits the following decomposition in bigrams: ta-ab-bl-le).

(2) Definition of orthographic neighbor:

A word $w_2$ is said to be an orthographic neighbor of word $w_1$ if and only if $w_2$ can be obtained simply by changing one of the letters of $w_2$. For instance, the word "cable" is an orthographic neighbor of "table". Similarly, "used" is an orthographic neighbor of "uses". Thus, given a generic word the process of computing its orthographic neighbors consists in the generation of all the possible permutations, using the target language alphabet, changing only one character at a time of the target word, and removing those permutations resulting in meaningless words.

The first application is based on a cognitive psychology experiment focused on determining the effect of word-syllable frequencies in a visual word recognition task in Spanish. In this task, both words and nonwords with very specific properties are needed. In particular, nonwords from two up to four syllables were generated using the following criterion: The average summation of the positional frequencies of the within-syllable bigrams must be less or equal than the average summation of the positional frequencies of the between-syllable bigrams. The positional frecuency of a bigram is computed as the number of occurrences a particular bigram appears in a specific position within a word (eg: The bigram "ic" appears in the fifth position within the word "pacific" and in the sixth position in word "specific"). For both experiments the language source database used was LEXESP [9]. In addition, the frequencies of the bigrams can be computed according to two different measures: token and type. Computing the token frequency implies using the entire database, while the type frequencies are computed using a subset of the whole language database that is generated by sorting the entire database and removing the occurrence of word repetitions.

The second application was focused on determining the robustness of the proposed algorithm in a more complex computational task consisting in generating nonwords of between two and four syllables, with a number of orthographic neighbors from one to four.

### 4.1. OBJECTIVE FUNCTION

According to the methodology presented in section 3, the first step to be performed is to transform the 0-1 criterion into a real-valued function. For the first experiment, we can express this fact by the following function:

$$C_b^1(w) = \begin{cases} 1 & \frac{1}{L-N}\sum_i^N \Psi_w(s_i) \leq \frac{1}{N-1}\sum_i^{N-1} \Psi_b(s_i, s_{i+1}) \\ 1 - \dfrac{\left|\dfrac{1}{L-N}\sum_i^N \Psi_w(s_i) - \dfrac{1}{N-1}\sum_i^{N-1} \Psi_b(s_i, s_{i+1})\right|}{\Delta_b} & \text{Otherwise} \end{cases}$$

The input to the previous function is a nonword segmented into its corresponding syllables. In the above expression the following notation is used:

$N$: Number of word-syllables

$L$: Length of the input nonword in characters

$s_i$: Syllable i of the input nonword $w = s_1 s_2 ... s_N$

$\psi_w(x)$: This function computes the positional frequencies of the within-syllable bigrams of syllable x.

$\psi_b(x,y)$: This function computes the positional frequencies of the in between-syllable bigrams. Specifically, between syllable x and syllable y.

$\Delta_b$: Normalization factor just to ensure function values remain in the real interval [0,1]. In our experiments was set to 74018.64 and 1399.09 for token and type frequencies respectively.

The objective function used for the orthographic neighbors experiment can be expressed by the following function:

$$C_n^1(w) = \begin{cases} 1 & \Psi_n(w) \in [a,b] \\ 1 - \dfrac{\left|\dfrac{a+b}{2} - \Psi_n(w)\right|}{\Delta_n} & \text{Otherwise} \end{cases}$$

In the above expression, the input to the function is a nonword $w$. The function $\Psi_n$ returns the number of orthographic neighbors of the input nonword. The constant $\Delta_n$ is a normalization factor equal to 100 in our experiments. The interval [a,b] represents the desired range of neighbors of the target nonwords.

### 4.2. PROBLEM REPRESENTATION

Following the notation introduced in section 2, our case study used as word unit $\mu$ the syllable. The cardinality $\lambda$ of this unit in Spanish is about 4000. Solutions were encoded as an array of integers, where each component of the array represents a word syllable. Specifically, each component indicates a position inside an array of strings where all the possible syllables are stored. The number of components of the solution vectors provides the number of syllables of the nonword. For example, a specific nonword of three syllables could be represented as the following permutation: S = [123 346 888]. The real nonword under the previous encoding can be obtained by concatenating the strings at the positions specified by the components of vector S when accessing the array containing all the possible syllables. Therefore, each component of any solution vector is bounded within the interval [0, $\lambda$-1]. It is important to note that the proposed encoding is universal and can be used for any other word unit (eg: the morpheme) of any given language.

### 4.3. NEIGHBORHOOD GENERATION

The process of neighborhood generation can be stated as follows. From the current configuration point $v$ an elementary move is performed by replacing one of the components of vector $v$, that is, $v(i)$ by a value obtained from a randomly generated set of points which are bounded by the cardinality of the word unit employed. This procedure is repeated in turn for each of the vector dimensions and using all the values contained in the random set.

```
function neighborhood_generation ( v )

i ← 0       (counter initialization)
repeat
    χ(i) ← random [0, λ-1]
    i ← i+1
until ( ‖χ‖ ≤ χ_max )
j ← 0 ; k ← 0; l ← 0;
for j in δ    (for each component of vector v)
    for k in χ
        w ← v(j)
        v(j) ← χ(k)
        M(l) ← v
        l ← l+1
        v(j) ← w
    endfor
endfor

return M

endfunction
```
Figure 4: Skeleton of the neighborhood generation procedure

Figure 4 depicts the process of neighborhood generation. The above procedure generates a Von Neumann-like neighborhood [10] from the current configuration point $v$. In particular, it generates local search directions which are normal to the hyperplanes defined by the cartesian axes. The data structure $\chi$ must ensure the absence of duplicates from the randomly generated points. The resulting procedure returns a $\chi_{max} \times \delta$ matrix containing the elementary moves to apply from the current configuration point.

## 5. EXPERIMENTAL RESULTS

The performance of the algorithm was tested for the applications exposed in section 4 by using a

```
function combinatorial_iterated_local_search

t ← 0                (iteration counter)
D^(0) ← ∅            (data structure to keep track of solutions found)

repeat
    v^(t) ← random   (Generate a random configuration point v)
    local_search(v)  (D^(t) is implicitly changed)
    t ← t+1

until ‖D‖ is acceptable or maximum number of iterations reached

endfunction
```
Figure 5: Skeleton of the adapted iterated local search algorithm

randomized local search algorithm. This kind of algorithm is usually included in the benchmarks of advanced optimization algorithms for comparison purposes [11]. Furthermore, these algorithms are, in fact, the simplest form of an iterated local search scheme [12] [13] . We adapted the above-mentioned algorithm to account for the combinatorial search task., denoting the modified algorithm as *Combinatorial Iterated Local Search* (CILS hereafter). In particular, it is based on the repeated generation of random configurations that are used as starting points for a local search algorithm. The pseudocode of the algorithm is shown in figure 5.

The local search procedure simply generates a neighborhood of the current solution $v$ by using the algorithm presented in the previous subsection. Thus, a more reliable measure of quality can be obtained when comparing both algorithms. Afterwards, the points of the neighborhood are evaluated using the functions described in section 4. The set of points that accomplish the optimality criterion ($C^1 = 1$) are inserted into the data structure $D$.

The results of averaging ten simulations for each of the algorithms are illustrated in tables 1, 2 for the first experiment and table 3 for the second. We ran each instance of the algorithms 500 iterations. The parameters used of the algorithm used for the simulations were *REP* = *CHAOS* = 3, *INCREASE* = 1.3, *DECREASE* = 0.8, $\chi_{max}$ = 300 and $R_{max}$ = 8000. Simulations were carried out on a workstation running Windows 2000 as operating System on a Pentium III 930 Mhz, 128 Mb of RAM.

| Combinatorial Reactive Search | Two Syllables | Three Syllables | Four Syllables |
|---|---|---|---|
| Solutions found | 24305 | 56431 | 59524 |
| Standard deviation | 2147.80 | 16635 | 15015 |
| Running time (s) | 16.90 | 28.87 | 39.40 |

| Combinatorial Iterated Local Search | Two Syllables | Three Syllables | Four Syllables |
|---|---|---|---|
| Solutions found | 17212 | 13659 | 4683 |
| Standard deviation | 595.68 | 864.85 | 636.33 |
| Running time (s) | 14.31 | 22.60 | 23.67 |

Table 1: Results for the CRS and CILS algorithms for the bigrams token frequency problem.

The algorithms were written in JAVA and compiled and tested using the JDK1.3.1. A major advantage of using an object-oriented language like JAVA is the flexibility it provides for re-use existing code and rapid prototyping capabilities. In this sense, nonword generation, as we have stated before, is subject to very difficult and changing criteria that depend on the particularities of the experiment or the application context. Therefore, the fact of using an object-oriented language permits the templatization of the nonword generation criterion by simply redefining certain steps of the algorithm (eg: simply by subclassing and re-implementation of a class method) without changing the algorithm structure.

| Combinatorial Reactive Search | Two Syllables | Three Syllables | Four Syllables |
|---|---|---|---|
| Solutions found | 24383 | 50813 | 64781 |
| Standard deviation | 1176.50 | 1050 | 25052 |
| Running time (s) | 17.09 | 29.13 | 39.19 |

| Combinatorial Iterated Local Search | Two Syllables | Three Syllables | Four Syllables |
|---|---|---|---|
| Solutions found | 19165 | 14474 | 4965 |
| Standard deviation | 670.94 | 1153.60 | 562.01 |
| Running time (s) | 14.47 | 22.80 | 23.78 |

Table 2: Results for the CRS and CILS algorithms for the bigrams type frequency problem.

The simulation results show that the CRS scheme outperforms CILS in all of the problem instances, although this is accomplished through a slight increase in the computation time. In addition, the running times for the orthographic neighbors problem (table 3) are one order of magnitude bigger than for the bigrams frequency problem due to the higher computational load introduced by this task. In general, the computational cost per iteration is greater in the CRS scheme than in CILS, nevertheless, this is not always the case as it depends on how often the algorithm enters into a diversification phase and also on its length.

| Combinatorial Reactive Search | Two Syllables | Three Syllables | Four Syllables |
|---|---|---|---|
| Solutions found | 15781 | 8492.40 | 4370 |
| Standard deviation | 643.32 | 1304.70 | 1678.10 |
| Running time (s) | 211.32 | 366.90 | 487.58 |

| Combinatorial Iterated Local Search | Two Syllables | Three Syllables | Four Syllables |
|---|---|---|---|
| Solutions found | 11135 | 55.40 | 0 |
| Standard deviation | 711.23 | 16.31 | 0 |
| Running time (s) | 153 | 186.31 | 93.84 |

Table 3: Results for the CRS and CILS algorithms for the orthographic neighbors problem.

Unlike CILS, which is strongly affected by the problem dimensionality, in the CRS scheme the effect is practically the opposite. More precisely, the proposed algorithm presents a strong robustness against problem dimensionality.

## 6. CONCLUSIONS

In this paper we have investigated the application of a meta-heuristic algorithm suitable for combinatorial optimization problems in a merely combinatorial search problem. Throughout this paper we have referred to the concept of combinatorial search as the problem of finding the highest amount of solutions matching a certain 0-1 criterion over a vast combinatorial space.

We have presented a formal description of the problem in terms of its application context. Specifically, within the Cognitive Neuroscience research field. We have also shown how to adapt the Reactive Search framework of algorithms to address a combinatorial search problem. In addition to the changes shown for the basic RS functions, several successive steps must also be performed in this regard:

- Relaxation of the 0-1 search criterion by transforming this constraint into a real-valued function, using a distance function to provide a measure of how far a solution is from the optimality criterion.

- Randomization of the neighborhood generation procedure to promote the exploration capabilities of the algorithm

The experimental results clearly show the algorithm is in fact able to generate nonwords of any size and subject to any criteria, since the proposed encoding scheme is universal. The abilities of this model suggest the applicability of the proposed methodology to other domains. Although further research must be carried out, one of the important conclusions of this work is that the reaction and feedback mechanisms introduced by this model offers a good alternative to classic random generation techniques that cannot cope adequately with a combinatorial search. Furthermore, they cannot offer general solutions to combinatorial search problems. Another interesting feature of the algorithm is its robustness against problem dimensionality.

Our current work is focused on extending the proposed method to cope with multi-objective combinatorial search. We contemplate also the possibility of including this algorithm in a more general software tool of use to the neuroscience community.